\documentclass{article}
\usepackage[utf8]{inputenc}
\usepackage[pdftex]{graphicx}
\usepackage{subcaption}
\usepackage{amsmath}
\usepackage{authblk}
\usepackage[symbol]{footmisc}

\title{Review of Factor Graphs for Robust GNSS Applications}

\author[1]{Shounak Das}
\author[2]{Ryan Watson}
\author[1]{Jason Gross}
\affil[1]{Department of Mechanical and Aerospace Engineering \\
West Virginia University \\
Morgantown, WV 26505, USA }
\affil[2]{The Johns Hopkins University Applied Physics Laboratory, Laurel, USA}
\date{}

\begin{document}

\maketitle

\section{What is a factor graph?}

Probabilistic modelling is a very important tool for any kind of estimation problem. This led to the development of Graphical Models \cite{koller2009probabilistic} which model known and unknown variables of a system using vertices in a graph and define edges between them to represent their relationship. Due to the inherent uncertainty in modelling a system, these relations are probabilistic. A factor graph connects many types of these graphical models like Markov Random Fields, Bayesian Networks, Tanner Graphs \cite{frey1997factor}. The primary motivation of a factor graph is to represent a global function of many variables as a product of local functions with smaller subsets of variables. The factor graph is not a method but a framework for modelling any system using its locality structure, that is each variable is only dependent on a few other local variables and is independent of the others. As explained in \cite{dellaert2021factor}, it is this locality property that makes it useful in modelling a variety of problems including mapping, visual-inertial odometry, motion planning, trajectory estimation and deep learning. The factor graph is defined as a Bipartite graph which has two types of vertices, one is the variable (i.e., the state vector) vertex which is to be estimated and another one is the factor vertex which encodes the constraints (e.g., a set of GNSS observations) applied to the variable vertices. An edge can only exist between a factor vertex and a variable vertex. The factor vertices represent the local functions which depend on the variable vertices with which it shares edges. A common estimation problem in robotics uses the factor graph framework to estimate the unknown robot pose along with other parameters depending on the problem. This is achieved by solving the Maximum-A-Posteriori (MAP) problem which maximizes the product of factors that are probabilistic constraints between states and measurements. \cite{dellaert2017factor} gives a rigorous mathematical description of how this maximization problem is solved in the field of robotic perception.

 Factor graphs have been treated as an alternative framework for solving estimation problems and have proven very effective for specific applications including Simultaneous Localization and Mapping (SLAM). This framework has attracted widespread use due to the flexibility of the approach and the ease of implementation granted through the availability of open-source graph optimization libraries like GTSAM \cite{dellaert2012factor}, g2o \cite{grisetti2011g2o}, Ceres \cite{ceres-solver}. While the factor graph framework has been shown to be beneficial for many applications, it should be noted that the frame work can be used in an equivalent manner to (i.e., it is a generalization of) existing state estimation implementations (e.g., the Kalman filter and it many variants). To begin this comparison, we will note that the factor graph ultimately encodes an objective function, which is solved through repeated relinearization via a non-linear optimization routine (e.g., Gauss-Newton, Levenberg–Marquardt). Previous work has demonstrated that iterating on the measurement update of Extended Kalman Filter and relinearizing the system models between iterations is equivalent to a Gauss-Newton optimization \cite{bell1993iterated}. Which shows that, under a certain set of constraints, the factor graph operating in batch mode is equivalent to a backward-smoothing Extended Kalman Filter (EKF) \cite{psiaki2005backward} that re-linearizes system and observation models and iterates.

\vspace{10 pt}
\section{What are the advantages of using a factor graph?}
Factor graph optimization has some advantages over the standard non-iterated Kalman filter which may be valuable for certain applications. First, like any optimization problem, it uses multiple iterations to minimize a cost instead of just one iteration for each state as in a standard Kalman filter. Second, it also linearizes the nonlinear measurement model every iteration step for every state unlike the single linearization performed by the standard Kalman filter. Factor graphs have also been shown to better exploit the time correlation between past and current epochs, which has been attributed to the batch nature of the estimation method. In particular, when operating in a batch-mode, a factor graph would be equivalent to a forward filter and backward smoother after each measurement update. For GNSS/INS applications, these benefits have been supported by experimental results in \cite{wen2021factor} where factor graphs have been shown to perform better than an EKF in urban environments. It might appear that with accumulation of new measurements over a longer time, the batch estimation might lose real-time performance. A sliding window approach can also be used similar to \cite{wilbers2019localization} to relieve computational cost.The window size has been found to be crucial for good optimization results and can depend upon environmental conditions \cite{wen2021factor}. Factor graphs achieve efficient computation by utilizing the sparse nature of the Jacobian and information matrices. This helps in fast matrix factorization and back-substitutions. Directly removing earlier poses from the graph can lead to information loss. This can be avoided using marginalization in the square-root information form which removes variables from the Bayes net derived from the factor graph using elimination algorithm. Due to the sparseness in the graph, incremental QR factorization can be also achieved efficiently \cite{dellaert2012factor}. Beyond fixed-lag smoothing, the \textit{isam2} formulation~\cite{kaess2012isam2} achieves real-time performance by converting the factor graph to the Bayes tree \cite{kaess2010bayes} when a new constraint is added. This is a more accurate incremental and smoothing method for highly nonlinear measurement models. The vertices of Bayes tree represent cliques in the Bayes net that is obtained from the factor graph during factorization. Only states contained within the same cliques as the states in the new constraint  and their predecessors in the Bayes tree need to be updated. Watson and Gross \cite{watson2018evaluation} used \textit{isam2} in a GNSS factor graph to show improved positioning performance than a traditional EKF-Precise Point Positioning (PPP) method. Wen et al. \cite{wen2021towards} applied factor graph optimization to the problem of both GNSS and GNSS-Real-time Kinematic (RTK) positioning and shows better performance than an EKF.

\section{What does a factor graph applied to GNSS look like?}

\begin{figure}[h]
\centering
\includegraphics[width=1.0\linewidth]{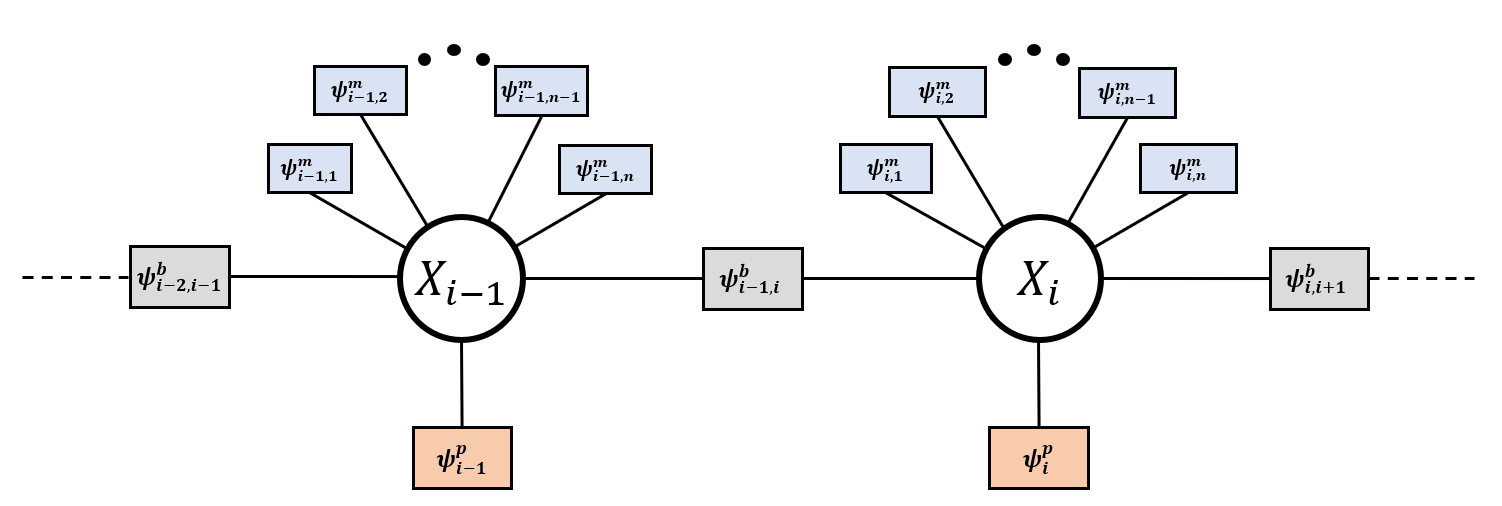}
\caption{GNSS factor graph example. }
\label{fig:factorgraph}
\end{figure}

A detailed description of creating factors with GNSS observations is presented in~\cite{watson2018evaluation}. The states commonly estimated in the GNSS factor graph are the receiver position, tropospheric delay, carrier phase bias, and the receiver clock bias. A visual representation of the GNSS factor graph is provided in Fig.~\ref{fig:factorgraph}, where $\psi$ represents any probabilistic constraint that might exist between the states and the measurements. In this specific implementation, $\psi^p$ encodes the prior belief on each state, which depends on the specific data set and environmental characteristics. $\psi^b$ is a motion constraint between two consecutive states along the trajectory which could, for example, incorporate motion data from an Inertial Measurement Unit (IMU) or wheel odometry. A common example of $\psi^b$ used in GNSS/IMU navigation is the factor which uses IMU-preintegration \cite{forster2015imu} to calculate displacement between the two factor graph locations with multiple IMU measurements integrated between them. Finally, $\psi^m$ are measurement constraint between a state and the measurements that were perceived from that state, for example GNSS pseudorange or carrier-phase measurements. To find the MAP estimate for the GNSS factor graph, we could find the set of states that maximize the product of factors. However, in practice, this optimization problem can be greatly simplified by employing the Gaussian noise assumption, which enables the conversion of the problem from maximizing the product of the factors to a non-linear least squares problem where each component of the sum is a Mahalanobis cost, which represents sum of squares of the normalized residuals, as provided in Eq. \ref{eq:factorcost}, where $f(*)$ a mapping between states at difference epochs and $h(*)$ is a mapping from the state space to the observation space. 

% \begin{equation} 
% \label{eq:factorprod}
% \hat{X}=\underset{x}{\operatorname{argmax}}\left\{\prod_{i=1}^{I} \psi_i^p \prod_{j=1}^{J} \psi_{j-1,j}^b \prod_{k=1}^{K} \psi_k^m\right\}
% \end{equation}

% \begin{equation}
% \label{eq:factorcost}
% \hat{X}=\underset{x}{\operatorname{argmin}}\left[\sum_{i=1}^{I}\left\|x_{o}-x_{i}\right\|_{\Sigma}^{2}+\sum_{j=1}^{J}\left\|x_{j}-f_{k}(x_{j-1})\right\|_{\Lambda}^{2}+\sum_{k=1}^{K}\left\|y_{k}-h\left(x_{k}\right)\right\|_{\Xi}^{2}\right]
% \end{equation}

\begin{equation}
\label{eq:factorcost}
\begin{split}
\hat{X} & =\underset{x}{\operatorname{argmin}}\left[\sum_{i=1}^{I}\left\|\psi^p_i\right\|_{\Sigma}^{2}+\sum_{j=1}^{J}\left\|\psi^b_j\right\|_{\Lambda}^{2}+\sum_{k=1}^{K}\left\|\psi^m_k\right\|_{\Xi}^{2}\right] \\ & =\underset{x}{\operatorname{argmin}}\left[\sum_{i=1}^{I}\left\|x_{o}-x_{i}\right\|_{\Sigma}^{2}+\sum_{j=1}^{J}\left\|x_{j}-f_{j}(x_{j-1})\right\|_{\Lambda}^{2}+\sum_{k=1}^{K}\left\|y_{k}-h_{k}\left(x_{k}\right)\right\|_{\Xi}^{2}\right] 
\end{split}
\end{equation}

\section{What are methods for robust estimation using factor graphs in GNSS?}
As mentioned, the factor graph framework also makes it easy to add existing and new robust estimation methods which can help reduce localization error during spoofing attacks or large noise from multipath or atmospheric effects. The discussion below enumerates some of these robust methods applied to GNSS.

Sunderhauf et al. \cite{sunderhauf2012switchable} defined Switch Constraints (SC), which is a lifted optimization \cite{zach2017iterated} methodology, that defines an observation weighting function $\Psi()$ that is a function of switch variables $s$, which is estimated in conjunction with the state parameters of interest. The SC method was initially developed for robust loop closure detection in SLAM and then extended to GNSS for multipath mitigation \cite{sunderhauf2012multipath}. When utilizing switch constraints, the pseudorange factor cost is expressed as a scaled version of the Mahalanobis cost between the predicted and actual measurement
\begin{equation}
    \left\|\mathbf{e}_{k}^{\mathrm{switch}}\right\|_{\boldsymbol{\Sigma}_{k}}^{2}=\left\|\Psi\left(s_{k}\right) \cdot\left(y_{k}-h_k(x_k)\right)\right\|_{\boldsymbol{\Sigma}_{k}}^{2}, 
\end{equation}
where the function $\Psi$ is a linear function of the switch variable. Prior factors are added for each of these switch variables to stop the optimization from making all $s_{k}$ to zero. A transition factor can also be added to model the change between $s_{k-1}$ and $s_{k}$ if the same satellite is observed at the next time step. These switch functions help in automatically de-weighting erroneous measurements (e.g., suspected multipath measurements) and are seen to perform better than computationally expensive ray tracing methods \cite{sunderhauf2012multipath}.

An extension of SC was derived in \cite{agarwal2013robust} called Dynamic Covariance Scaling (DCS) where the switch variables are taken out of the optimization method and calculated separately using the residual, current measurement uncertainty and a prior switch uncertainty. After calculating $s_{k}$, the information matrix associated with the GNSS observation factor is scaled by $\Psi(s_{k})^2$.

Max-mixtures (MM) \cite{olson2013inference} was also developed to tackle false loop closures using a Gaussian Mixture Model (GMM) but instead of the $sum$ operator which is unsuitable for maximum likelihood when multi-modal uncertainty model is utilized, the objective function is converted to use the $max$ operator, as shown in Eq. \ref{eq:max-mix}.

\begin{equation}
    \label{eq:max-mix}
    p\left(y_{i} \mid x\right)=\max _{k} w_{k} N\left(\mu_{k}, \Lambda_{k}^{-1}\right)
\end{equation}

The benefits of SC, DSC, and MM have been evaluated in \cite{pfeifer2016robust,watson2017robust} for GNSS factor graph applications with real world data. Both of these studies showed the substantial positioning improvement that can be granted via the utilization of robust estimation techniques when conducting optimization with degraded GNSS observations.

To extend upon the max-mixtures work, Watson et al. \cite{watson2019enabling} proposed to learn the GMM during run-time based upon clustering of the observation residuals. Initially, this work was implemented in a batch framework; however, it was later extended to work incrementally \cite{watson2020robust}, through an efficient methodology for incrementally merging GMMs.

M-estimators \cite{bosse2016robust} have also been recently tested within the GNSS framework in batch form \cite{crespillo2020design} and found to perform better than non-robust estimators. M-estimators assume a loss function that is different from the squared loss function. The squared loss function is highly sensitive to outliers since it grows aggressively for larger values of residuals. Thus, a group of loss functions were introduced which grow less aggressively than the squared loss function. The Huber cost function \cite{huber1992robust}, as provided in Eq. \ref{eq:huber} is one such function.

\begin{equation}
    \label{eq:huber}
    \rho(z)=\begin{cases}z^{2} / 2 & |z| \leq \Delta \\ \Delta|z|-\Delta^{2} / 2 & |z|>\Delta \end{cases}
\end{equation}
When the objective function is modified to utilize a m-estimator, the optimization problem takes a form as depicted in Eq. \ref{eq:irls}. 
\begin{equation}
\label{eq:irls}
    \hat{X}=\underset{x}{\arg \min } \sum_{i} \rho\left(\frac{r_{i}(x)}{\sigma}\right),
\end{equation}
where $r_{i}(x)$ is the residual for each measurement and $\sigma$ is the scale parameter. 

Increasing the $\Delta$ parameter makes this function closer to the squared loss function. Equation~\ref{eq:irls} can be solved iteratively with weighted least squares method \cite{crespillo2020design, bosse2016robust}. Selecting a suitable $\Delta$ parameter is not straightforward, since it depends on the measurement noise statistics. Agamennoni et al. \cite{agamennoni2015self} uses the fact that some M-estimators like Huber, Cauchy, Laplace have a corresponding elliptical distribution to estimate the $\Delta$ and the states in an Expectation Maximization (EM) framework. Barron \cite{barron2019general} jointly optimizes for the states and the parameters for computer vision applications.
A factor graph gives greater flexibility in the M-estimator application since it can help in de-weighting not only the current measurements but also changing the weights of the past measurements. It also can help is totally removing some past measurements if it is found to be an outlier later whereas in the Kalman filter, the contribution of past measurements cannot be changed in a real-time manner. Most graph optimization libraries also have built in functionality to use robust cost functions which is also helpful.

Finally, the robust estimation technique derived by Yang et al. \cite{yang2020graduated} unites the two well-known ideas from computer vision, the Black-Rangarajan Duality \cite{black1996unification} and Graduated Non-Convexity \cite{blake1987visual} to iteratively solve the point cloud registration problem using robust cost functions. According to the Black-Rangarajan Duality, equation ~\ref{eq:irls} can be re-written as 

\begin{equation}
 \hat{X}, \mathbf{w}=\underset{x,w_{i} \in[0,1]}{\arg \min }\sum_{i}\left[w_{i} r_i^{2}\left(x\right)+\Phi_{\rho}\left(w_{i}\right)\right],
\end{equation}
where $w_i$ is the weight for the $i^{th}$ measurement and $\Phi_{\rho}$ is a penalty term which depends on the weight and the robust cost $\rho$. Graduated Non-Convexity(GNC) is a method to minimize a non-convex function $f$ without facing the problem of local minima. The idea is to replace the function $f$ with a surrogate function $f_{\mu}$ whose convexity is controlled by the parameter $\mu$. The optimization starts off with a convex form of $f_{\mu}$ and $\mu$ is changed iteratively such that the non-convexity increases. By uniting these two methods, \cite{yang2020graduated} solved two problems in equation~\ref{eq:irls}, 1) avoiding local minima while optimizing a robust cost function 2) convert equation~\ref{eq:irls} to a weighted least squares problem which is always easier to solve. The similarity of this problem to GNSS should be obvious to the reader. \cite{wen2021gnss} shows importance of applying this robust estimation technique in GNSS factor graph in mitigating multipath effects in urban canyons. The batch nature of most of these robust estimation methods makes it suitable for use in a factor graph rather than an EKF.

\section{What are potential uses of factor graphs for the GNSS and radio-navigation community?}
The EKF has been the preferred choice for GNSS based state estimation  due to their simplicity, computational efficiency, and the fact that GNSS observation models are well modeled by linear approximations and are often well characterized by Gaussian errors. Despite this, there may be situations in which the radio-navigation community can benefit from factor graph optimization. 

For one, recent work has emphasized the potential use and benefits of signals of opportunity (SOP) in radio-navigation applications. SOP may include as cell phone signals \cite{shamaei2021receiver, shamaei2018lte,morales2019orbit} and Low Earth Orbiting satellites. Use of SOPs often includes a need to solve for an unknown or very uncertain transmitter location and clock offsets. Therefore, this class of problems shares many parallels with SLAM and therefore may enjoy similar benefits from the use of a factor graph as recognized for visual or LIDAR based pose-graph SLAM.

Next, as discussed in the context of GNSS, many robust estimation techniques have been developed for use in factor graphs. For use of GNSS urban environments that is prone to multipath errors, the use of these robust factor graphs may be beneficial. For example, the winning solution of the Google SmartPhone decimeter challenge, which included a variety of datasets collected different environmental settings, was indeed a factor graph implementation \cite{googleChallenge2021}.

Finally, factor graph adoption may be beneficial simply because the factor graph framework has become the standard state estimation paradigm within the robotics and autonomy communities (i.e., pretty much every sensor modality, other than GNSS, utilizes -- and has shown the benefit of -- the factor graph framework). The adoption of a GNSS factor graph may enable more seamless integration between GNSS and other sensor modalities and integration of multiple information sources is a well-recognized key to any critical navigation system.

\bibliographystyle{IEEEtran}
\bibliography{reference}
\end{document}